\documentclass[journal]{IEEEtranTIE}
\usepackage{graphicx}
\usepackage{cite}
\usepackage{picinpar}
\usepackage{amsmath}
\usepackage{url}
\usepackage{flushend}
\usepackage[latin1]{inputenc}
\usepackage{colortbl}
\usepackage{soul}
\usepackage{multirow}
\usepackage{pifont}
\usepackage{color}
\usepackage{alltt}
\usepackage[hidelinks]{hyperref}
\usepackage{enumerate}
\usepackage{siunitx}
\usepackage{breakurl}
\usepackage{epstopdf}
\usepackage{pbox}
\usepackage{textcomp}
\usepackage{stfloats}
\usepackage{url}
\usepackage{verbatim}
\usepackage{graphicx}
\usepackage{cite}
\usepackage{tabularx}
\usepackage{booktabs}
\usepackage{epstopdf}
\usepackage{epsfig}
\usepackage{algorithm}
\usepackage{algorithmic}
\usepackage{multirow}
\usepackage{makecell}
\usepackage{amsmath,amsfonts}
\usepackage{algorithmic}
\usepackage{algorithm}
\usepackage{array}
\DeclareMathOperator*{\argminA}{arg\,min} % Jan Hlavacek
\begin{document}
\title{	Co-training partial domain adaptation networks for industrial Fault Diagnosis}

\author{
	\vskip 1em
	
	Gecheng Chen

}

\maketitle
	
\begin{abstract}
The partial domain adaptation (PDA) challenge is a prevalent issue in industrial fault diagnosis. Current PDA approaches primarily rely on adversarial learning for domain adaptation and use reweighting strategies to exclude source samples deemed outliers. However, the transferability of features diminishes from general feature extraction layers to higher task-specific layers in adversarial learning-based adaptation modules, leading to significant negative transfer in PDA settings. We term this issue the adaptation-discrimination paradox (ADP). Furthermore, reweighting strategies often suffer from unreliable pseudo-labels, compromising their effectiveness. Drawing inspiration from traditional classification settings where such partial challenge is not a concern, we propose a novel PDA framework called Interactive Residual Domain Adaptation Networks (IRDAN), which introduces domain-wise models for each domain to provide a new perspective for the PDA challenge. Each domain-wise model is equipped with a residual domain adaptation (RDA) block to mitigate the ADP problem. Additionally, we introduce a confident information flow via an interactive learning strategy, training the modules of IRDAN sequentially to avoid cross-interference. We also establish a reliable stopping criterion for selecting the best-performing model, ensuring practical usability in real-world applications. Experiments have demonstrated the superior performance of the proposed IRDAN.

\end{abstract}


%\markboth{IEEE TRANSACTIONS ON INDUSTRIAL ELECTRONICS}%
{}

\definecolor{limegreen}{rgb}{0.2, 0.8, 0.2}
\definecolor{forestgreen}{rgb}{0.13, 0.55, 0.13}
\definecolor{greenhtml}{rgb}{0.0, 0.5, 0.0}

\section{Introduction}
\IEEEPARstart{F}{ault} 
diagnosis is crucial for ensuring the reliability, safety, and efficiency of complex industrial processes or facilities. As one of the most popular methods, Data-driven fault diagnosis leverages machine learning or statistical methods to detect and classify faults in industrial systems by analyzing large volumes of sensor data and operational records, enabling more accurate and efficient maintenance decisions \cite{song2018fault,wan2017manufacturing}. Given the strong feature extraction ability and end-to-end structures of deep learning (DL) techniques, such as fully connected neural networks (FNN), and convolutional neural networks (CNN), DL-based models have been widely applied to data-driven diagnosis and achieved distinguished performance \cite{yang2022paradigm,wu2018deep}. 

Traditional DL approaches only work well under the identical distribution assumption, i.e., the training and testing sets should have the same distribution. However, this assumption can hardly hold due to
changeable operating conditions, equipment wear, and environment noise \cite{li2018cross}. Deep transfer learning (DTL) provides a potential framework to deal with this distribution discrepancy problem by leveraging knowledge from related source domains to improve the learning of the target tasks \cite{chen2023transfer}. As the subfield of DTL, unsupervised domain adaptation (UDA) refers to the case where there are labeled source samples available but all the target samples are unlabeled. In this work, we mainly focus on the UDA problem in the industrial processes.

Even UDA-related approaches have yielded promising performance, they all assume the source and the target domains share the same label space, i.e., fault categories. However, we often obtain source
data from various machinery conditions for training as the source domain, but it is often difficult to collect target data for all potential categories in advance. As a result, the label space of the target domain is just a subset of the source domain. Under these scenarios, traditional UDP approaches may result in severe negative transfer \cite{cao2018partial}. To address this issue, researchers have increasingly focused on partial domain adaptation (PDA) approaches, which have been widely applied in industrial fault diagnosis tasks \cite{zhang2018importance,jiao2019classifier,li2020novel,li2020deep,qian2023cross,li2024principal,li2023instance}. to show the superiority of the proposed method. Most PDA methods follow a common two-step process to exclude irrelevant source samples and facilitate knowledge transfer: (a) training an adversarial learning module to align the source and target domains by mapping them to a new ``middle domain," and (b) selecting source samples related to the target domain using pseudo labels. However, this framework often lacks reliability and efficiency. On the one hand, the self-training procedure is prone to error accumulation due to the unreliability of pseudo labels, particularly in the early stages of training. Furthermore, the reduced number of source samples after excluding outliers may not be sufficient for learning domain-invariant features \cite{sahoo2023select}.  On the other hand, in the adversarial domain adaptation training, the feature transferability drops from general feature extraction layers to higher task-specific layers \cite{li2020deep}. In other words, there exists severe interference between the complex task-specific layers and domain adaptation layers. 
%As a result, the transformed source and target domain are not guaranteed to maintain their original discriminative features in the ``middle domain" given the complicated adversarial training process, and thus cause negative transfer. 
We refer to this dilemma as \textit{adaptation-discrimination paradox} (ADP). This dilemma can be more severe for PDA settings due to the influence of outliers from the source domain.  

%The "adaptation-discrimination paradox" is mainly due to task-oriented transformation and the domain adaptation-oriented transformation will hurt each other. 
The intuition of this work comes from the basic classification task with the identical distribution assumption. Users and researchers do not consider any ``partial strategies" when the label space of the testing set is a subset of the training set. On the contrary, it has been proven that the training set with more categories (e.g., ImageNet) can help learn the relationship between different classes and thus generalize the testing set better.  
Building on this insight, we propose that if classifiers are constructed separately for each domain (i.e., \textit{domain-wise classifiers}) rather than for a combined "middle domain," this approach can effectively eliminate the partial domain challenge.
%First, the transformed two domains are not close enough due to the equilibrium problem of adversarial learning-based DA approaches (underfitting). Second, the shared transformation damages the discriminative features of the two domains because the domains with different label spaces will interfere with each other during the training. 
%the setting of the PDA should get as good performance as the close-set counterpart. Based on this, we argue that the negative transfer in PDA setting is mainly due to 1) happens in the feature transformation module. If 

Inspired by this, we construct a novel PDA framework called the interactive residual domain adaptation networks (IRDAN). We implement residual domain adaptation (RDA) blocks to construct domain-wise classifiers SEPARATELY. Each RDA block performs one-directional feature adaptation (i.e., from source to target or target to source). Subsequently, each classifier is trained using both its own domain's original samples and the transformed samples from the other domain via the RDA block. By following this approach, the discriminative features of the original (untransformed) samples are preserved, making them more reliable in guiding the training of task-specific modules. Furthermore, the transformation process is simpler and less prone to ADP problem, as it only handles one-directional adaptation. This new PDA framework addresses the PDA challenge from a different perspective, making full use of the available information without down-weighting or excluding any samples.

To prevent interference between feature transformation modules and task-specific (classification) modules, we introduce an interactive learning strategy that trains the modules of each domain-wise classifier sequentially, ensuring a confident information flow. In this process, the information from one domain is used to guide the training of the other, and the two classifiers are updated interactively, mitigating the risk of error accumulation commonly seen in self-training approaches. Additionally, we establish a reliable stopping criterion for this interactive learning process based on the agreement between the two classifiers. Unlike existing methods that rely on classifier inconsistency, such as Maximum Classifier Discrepancy (MCD) \cite{saito2018maximum}, our two classifiers are trained on entirely different features yet converge on consistent final decisions. Moreover, since our domain-wise classifiers are completely independent, we avoid adversarial training, thereby eliminating the adaptation-discrimination paradox (ADP) issue.

%More importantly, since we do not make transformations on the original samples of one domain for each classifier, their discriminative features are intact and thus reliable to guide the training and avoid negative transfer under PDA setting. What's more, t
%This new PDA framework does not down-weight or rule out any samples to make use of the provided information as much as possible. Compared with the existing bi-classifier domain adaptation methods, these two classifiers are fed absolutely different features in our framework but agree on the final decisions.

In summary, We highlight the main contributions of this work:
\begin{itemize}
\item We propose a novel framework featuring domain-wise classifiers to address the partial domain adaptation challenge.
\item The domain-wise classifiers are built with a residual structure to overcome the prevalent ``adaptation-discriminative paradox" in domain adaptation processes.
\item We develop interactive training strategies to sequentially train each component of the classifiers, preventing interference between them.
\item We establish reliable stopping criteria for the training process and design an effective decision-making strategy for the final model output.
\end{itemize}

%Existing PDA approaches mainly train a domain adaptation module (including a feature generator, a domain discriminator and a classifier) and select the related source samples based on pseudo labels given by the classifier. By implementing these two steps iteratively, the researchers believe the model can rule out the outlier categories and maintain an efficient knowledge transfer. 

\section{Interactive residual domain adaptation network}

\subsection{Problem formulation}
In this work, we focus on the fault diagnosis tasks under the unsupervised domain adaptation (UDA) setting, i.e., no labeled target samples available for training. Assume the labeled source domain $D_s=\{(x^i_s,y^i_s)\}_{i=1}^{n_s}$ and the unlabeled target domain $D_t=\{x^i_t\}_{i=1}^{n_t}$, where $n_s$ and $n_t$ are the sample size of the source and the target domain, respectively, $y^i_s \in \{1,..., K\}$ represents the fault categories. Our ultimate goal is to train a network for the task of cross-domain fault diagnosis, i.e., find accurate labels $y^i_t$'s for the target samples where the conditional and/or marginal distributions of these two domains are different, i.e., $P(x_s) \not= P(x_t)$ and/or $P_s(y_s|x_s) \not= P_t(y_s|x_s)$.

\subsection{Interactive residual domain adaptation network}

\begin{figure}[ht]
    \centering
    \includegraphics[width=0.5\textwidth]{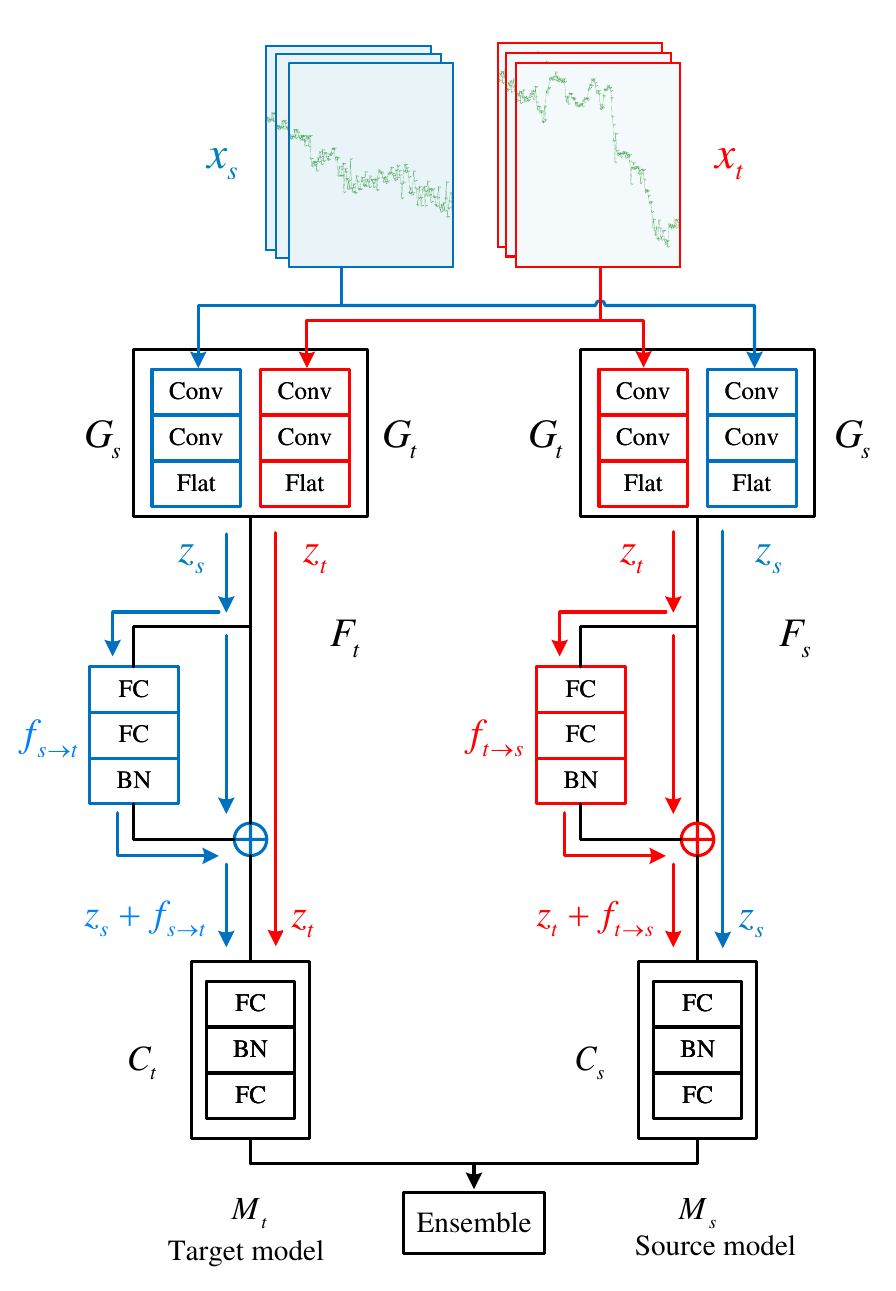}
    \caption{The structure of IRDAN.}
    \label{fig:structure}
\end{figure}

In this section, we introduce the structure of the Interactive residual domain adaptation network (IRDAN). The main idea of IRDAN is to construct domain-wise models separately, i.e., source model $M_s$ and $M_t$. These two models do not share any parts during the whole training process and the final diagnosis decision is based on an ensemble strategy. We use Figure \ref{fig:structure} to show the overall structure of the proposed method. We use lines in blue and red to show the flow of source and target samples (features), respectively. 

\subsubsection{General feature extractor based on contrastive learning}
Dynamic features are of great importance for the fault diagnosis in the complex industrial processes. As a result, we first follow the time-frequency transform as in \cite{shao2018highly} to convert the one-dimensional raw signals into two-dimensional images to better represent the dynamic features.

Then we construct a contrastive learning-based feature extractor as the beginning of the proposed structure. Contrastive learning aims to learn effective embeddings for images so that similar samples are pulled closer and diverse samples are pushed away \cite{bachman2019learning}. By doing this, the contrastive learning module can capture robust discriminative representations. We implement the SimCLR \cite{chen2020simple} to train the general feature extractors $G_s$ and $G_t$.

Denote input sets as $X_s=\{x^i_s\}_{i=1}^{n_s}$ and $X_t=\{x^i_t\}_{i=1}^{n_t}$. Also, denote $\hat{X}_s$ and $\hat{X}_t$ as the augmented version of $X_s$ and $X_t$, respectively. We implement the SimCLR separately to train $G_s$ and $G_t$. As for the feature extractor $G_s$, we consider the $X_s$ and $\hat{X}_s$ as the training sets. For each augmented sample $\hat{x}^i_s$, only the original sample $x^i_s$ is regarded as the positive pair, and all the other samples in $X_s$ and $\hat{X}_s$ are regarded as the negative pairs. We try to minimize the Normalized Temperature-Scaled Cross-Entropy loss (NT-Xent) defined as:
\begin{equation}\label{ntloss_s}
    \mathcal{L}_s^{con}=\frac{1}{N_s}\sum_{i=1}^{N_s}-\log \frac{d(v_i,v_i^+)}{d(v_i,v_i^+) + \sum_{v_i^-\in V_s^-}d(v_i,v_i^-)},  
\end{equation}
where $v_i,v_i^+$ represents the projections of positive pairs, $v_i,v_i^-$ represents projections of negative pairs and $V_s^-$ means the set of projections of negative pairs. $d(\cdot,\cdot)=\exp (sim(\cdot,\cdot)/\tau)$ calculates the similarity between two projections, $sim(\cdot,\cdot)$ represents a kernel function and $\tau$ is the temperature parameter.

As for the feature extractor $G_t$, we use $X_s$ and $\hat{X}_s$ as the training sets. Similar to $\mathcal{L}_s^{con}$, we have the NT-Xent for $G_t$ defined as:
\begin{equation}\label{ntloss_t}
    \mathcal{L}_t^{con}=\frac{1}{N_t}\sum_{i=1}^{N_t}-\log \frac{d(v_i,v_i^+)}{d(v_i,v_i^+) + \sum_{v_i^-\in V_t^-}d(v_i,v_i^-)}. 
\end{equation}
In each domain-wise model, we use $G_s$ and $G_t$ to process the source and target inputs, respectively. Denote the extracted features of source and target samples as 
\begin{equation*}
    z_s=G_s(x_s),
\end{equation*}
and 
\begin{equation*}
    z_t=G_t(x_t).
\end{equation*}

\subsubsection{Residual domain adaptation block}
%Existing DA approaches always need a complex adaptation module because they try to align the features from both domains in a "middle domain". Since many consecutive layers of non-linear transformations will amplify the feature distribution difference across domains and vice versa. As a result, this complex module can be highly possible to destroy the discriminative features of source or target samples, i.e., trigger the ADP problem.

%We refer to this dilemma as the "adaptation-discrimination paradox". Next, they can not align the source and target domains completely due to some systematic problem, such as the equilibrium of DANN-related structure. We refer to this problem as the "vanishing adaptation gradient" because it is highly similar to the vanishing gradient problem in traditional NNs. 

%because either they destroy the discriminative features related to the classification tasks (Catastrophic Forgetting) or they do not align the source and target domains effectively (Vanishing Gradients). To be specific, the main reasons for these two dilemmas can be summarized as 1) destroyed discriminative features of source samples; 2) destroyed discriminative features of target samples; 3) ineffective feature alignment.  

The residual domain adaptation block (RDA) implements one-direction domain adaptation to avoid ADP problem. Specifically, we keep the extracted features corresponding to its own domain unchanged and feed the extracted features from the other domain to the transformed module. The RDA block has two channels: an identical channel and a residual correction channel. Let's take $F_t$ in $M_t$ as an example to introduce the function of the RDA block. For the extracted feature $z_s$ and $z_t$ from the previous module, $z_t$ will pass the identical channel to keep the discriminative features while $z_s$ also passes the residual correction channel to learn the domain discrepancy $f_{s\xrightarrow{}t}$. We expect the distribution of modified source feature $z_s + f_{s\xrightarrow{}t}$ to be similar to target samples. To summarize, the output of $F_t$ of the source and the target features are as follows:
\begin{equation*}
    F_t(z_s) = z_s + f_{s\xrightarrow{}t},
\end{equation*}
\begin{equation*}
    F_t(z_t) = z_t.
\end{equation*}
Similarly, we get the outputs of $F_s$ as 
\begin{equation*}
    F_s(z_t) = z_t + f_{t\xrightarrow{}s},
\end{equation*}
\begin{equation*}
    F_s(z_s) = z_s.
\end{equation*}

This residual-based adaptation module holds great potential to solve the ADP problem. On the one hand, this module only implements source-to-target or target-to-source adaptation. This one-direction transformation is much easier to learn and needs a less complicated structure compared with the existing DA approaches. This lightweight structure can maintain the discriminative features better. On the other hand, the original features from two domains $z_s$ and $z_t$ maintain their impact on the training via identity mapping structure. By doing this, the interference of adaptation layers on task-specific layers can be eliminated. 

%avoids the "vanishing gradient" during the training of the adaptation module and      

%This residual-based adaptation module is much easier to learn and the discriminative features in the unchanged features remain intact. 

%First, the reason for the Catastrophic Forgetting in the UDA setting is mainly due to the destroyed discriminative features of source samples and/or target samples because the existing DA approaches try to align the features from both domains in a "middle domain".

%This residual adaptation  
%Specifically, we only implement one-direction adaptation for features $z_s=G_s(x_s)$ ($z_t=G_t(x_t)$) from $D_s$ ($D_t$) in $M_t$ ($M_s$) and keep the features $z_t=G_t(x_t)$ ($z_s=G_s(x_s)$) from $D_t$ ($D_s$) unchanged. $z=G(x)$  

%Please note that the target model $M_t$ is constructed for samples in the target domain, as a result, the features extracted from the target samples can be directly

%feed the extracted features $z=G(x)$ from its own domain directly to the classifier for each model.   

\subsubsection{General Feature Alignment}
In order to guide the training of RDA blocks and reach effective domain adaptation, we implement a general feature alignment model after the residual adaptation block via Maximum Mean Discrepancy (MMD) \cite{gretton2012kernel} in each domain-wise model. MMD is commonly used to quantify the similarity of the distributions of source and target features. In our framework, the features from the RDA block should lie in the same domain in each domain-wise model. Let's take the target model $M_t$ as an example. According to our assumption, $M_t$ is designed for the distribution of original target samples. As a result, the corrected features out from $F_t$ should be aligned to the target domain before feeding to the classifier $C_t$. In other words, we need to minimize the MMD between two distributions $P_t^s=P(F_t(z_s))$ and $P_t^t = P(F_t(z_t))$ as defined
\begin{equation}\label{mmdt}
    MMD_t(P_t^s, P_t^t)=\sup_{f \in \mathcal{F}}(\mathbb{E}_{a\in P_t^s}[f(a)] - \mathbb{E}_{a\in P_t^t}[f(a)]),
\end{equation}
where $\mathcal{F}$ represents the universal class of functions. Similarly, we also need to minimize the MMD between two distributions $P_s^s=P(F_s(z_s))$ and $P_s^t = P(F_s(z_t))$ for the source model $M_s$ as defined:
\begin{equation}\label{mmds}
    MMD_s(P_s^s, P_s^t)=\sup_{f \in \mathcal{F}}(\mathbb{E}_{a\in P_s^s}[f(a)] - \mathbb{E}_{a\in P_s^t}[f(a)]).
\end{equation}

\subsubsection{Domain-wise models fusion}

The final block of the domain-wise model is the classifier module $C_t$ or $C_s$, which takes the output of the RDA blocks to make a decision. We take the $C_t$ of the target model as an example to illustrate the decision-making process. Denote the logits of $C_t$ for source features $F_t(z_s)$ and target features $F_t(z_t)$ as
\begin{equation}\label{logitsts}
    c_t^s = C_t(F_t(z_s)),
\end{equation}
and
\begin{equation}\label{logitstt}
    c_t^t = C_t(F_t(z_t)).
\end{equation}
Also, denote the corresponding predicted label of the target model as
\begin{equation*}
    M_t(x_s) = softmax(c_t^s),
\end{equation*}
and
\begin{equation*}
    M_t(x_t) = softmax(c_t^t).
\end{equation*}
Similarly, we have the logits of $C_s$ and the corresponding predicted labels of the source model as 
\begin{equation}\label{logitsss}
    c_s^s = C_s(F_s(z_s)),
\end{equation}
and
\begin{equation}\label{logitsst}
    c_s^t = C_s(F_s(z_t)).
\end{equation}
Also, denote the corresponding predicted label of the target model as
\begin{equation*}
    M_s(x_s) = softmax(c_s^s),
\end{equation*}
and
\begin{equation*}
    M_s(x_t) = softmax(c_s^t).
\end{equation*}

The goal for each domain-wise model is to classify both the source and target samples correctly, even though they have different structures and work on different domains. To be specific, both domain-wise models should reach the same decision for each source and target sample as long as the training process converges. To make the framework more robust, we ensemble these two models by taking the average of logits to give the final decision, i.e., the final prediction of IRDAN for a target sample is defined as
\begin{equation*}
    M(x_t) = softmax(c_s^t + c_t^t).
\end{equation*}

%.However, different from the existing classifier inconsistency-based domain adaptation approaches, we do not use the agreement between 

\section{Interactive learning strategy for IRDAN}

%Existing classifier inconsistency-based domain adaptation approaches, which generally share certain parts of the model (e.g., feature extractor), are trained via an adversarial manner. Given this, basically, they are different versions of DANN by replacing the original task-related loss with the classifier inconsistency loss. As a result, they also inherit the "adaptation-discrimination paradox" and "vanishing adaptation gradient" problems. 

%The "adaptation-discrimination paradox" is mainly due to task-oriented transformation and the domain adaptation-oriented transformation will hurt each other. In the structure of IRDAN, the residual adaptation blocks avoid complex structure of domain adaptation mudoles. 
%Compared with them, our two domain-wise models are constructed separately for each domain and even have different structures to achieve reliable feature alignment and classification. 

\begin{figure*}[ht]
    \centering
    \includegraphics[width=0.9\textwidth]{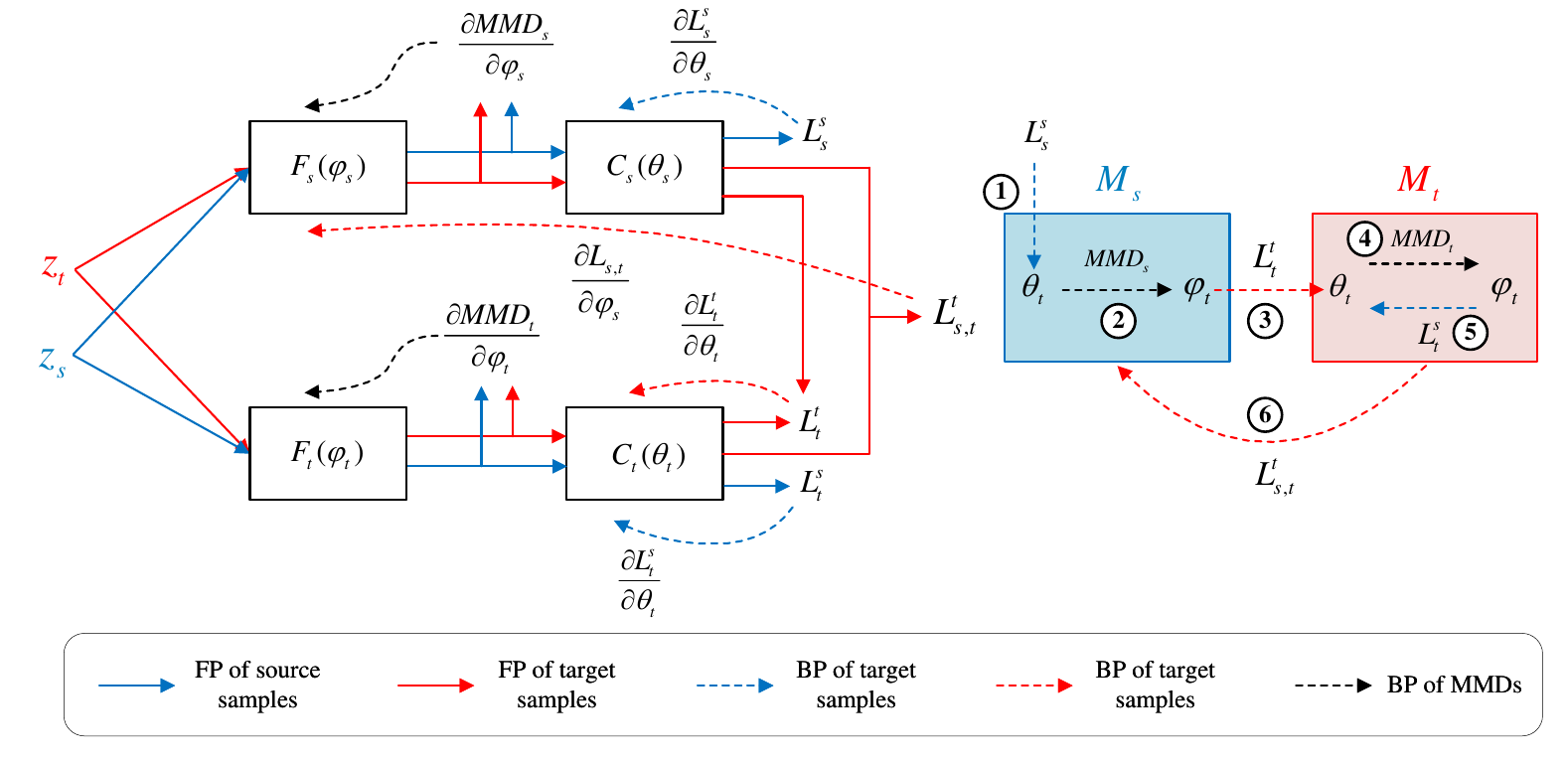}
    \caption{The interactive learning strategy for IRDAN. The left figure shows the forward (solid lines) and back (dashed lines) propagation caused by samples from source (blue) and target (red) domains and the black dashed lines represent the MMDs. The right figure shows the general step of the interactive training from step 1 to step 6.}
    \label{fig:loss}
\end{figure*}

In this section, we introduce the interactive learning strategy and the detailed steps for IRDAN. In general, we train different modules (general feature extractor $G$'s, residual adaptation blocks $F$'s, and classifiers $C$'s) of two domain-wise models separately to avoid interference between them. Figure \ref{fig:loss} shows the detailed training process of our method. The left figure in \ref{fig:loss} shows the forward (solid lines) and back (dashed lines) propagation caused by samples from source (blue) and target (red) domains and the black dashed lines represent the MMD's. The right figure in \ref{fig:loss} shows the general step of the interactive training from step 1 to step 6.

\subsection{Pre-training for $G_s$ and $G_t$}
First, we pre-train the general feature extractors for each domain, i.e., $G_s$ and $G_t$, through the contrastive learning. Let's denote the parameters of $G_s$ and $G_t$ as $\epsilon_s$ and $\epsilon_t$, respectively, then we implement the pre-training for $G_s$ and $G_t$ as
\begin{equation}
    \min_{\epsilon_s} \ \mathcal{L}_s^{con},
\end{equation}
and
\begin{equation}
    \min_{\epsilon_t} \ \mathcal{L}_t^{con}.
\end{equation}
Please note that these two extractors are trained separately for samples from two domains and we will freeze them in the following steps. Next, we need to train $F_t(\phi_t)$ and $C_t(\theta_t)$ of $M_t$ and $F_s(\phi_s)$ and $C_s(\theta_s)$ of $M_s$.

\subsection{Interactive training steps}\label{interactivelearning}
\subsubsection{Supervised training of $C_s$}
Let's begin with the source model $M_s$ because we have reliable label information for the source samples. We feed the source samples to the source model and train the classifier $C_s$ based on the label information. Denote the logits of $x_s^i$ given by $C_s$ as $c_s^{si}$ based on Equation \ref{logitsss}. We define the loss in this supervised learning as source-model source-sample loss $L_s^s$ as follows: 
\begin{equation*}
    \mathcal{L}_s^s = \sum_{i=1} ^{n_s} l(c_s^{si}, y_s^i),
\end{equation*}
and we implement the training
\begin{equation*}
    \min_{\theta_s} \ \mathcal{L}_s^s,
\end{equation*}
where $l(\cdot,\cdot)$ represents the cross-entropy loss. Note that the residual adaptation block $F_s$ provides an identity channel for the source samples, i.e., $c_s^{si} = C_s(z_s^{si})$, so we can train $C_s$ accurately to maintain the discriminative features. 

\subsubsection{Adaptation training of $F_s$}
Next, in order to align the target features to the source domain to match the working condition of the source model, we implement the general feature alignment to train the residual adaptation block $F_s$, i.e., minimize the $MMD_s$ defined in Equation \ref{mmds}. In other words, we implement the training
\begin{equation*}
    \min_{\phi_s} \ MMD_s.
\end{equation*}

\subsubsection{Supervised training of $C_t$ guided by $M_s$}
After these two steps, the source model should have certain confidence in the target labels which can be utilized as the basic information to guide the training of classifier $C_t$ of $M_t$. In other words, we use the output of $C_s$ to guide the training of raw $C_t$ in a supervised manner. Denote the logits of $x_t^i$ given by $C_s$ and $C_t$ as $c_s^{ti}$ and $c_t^{ti}$, respectively, based on Equation \ref{logitsts} and \ref{logitstt}. We define the target-model target-sample as 
\begin{equation*}
    \mathcal{L}_t^t = \sum_{i=1} ^{n_t} l(c_t^{ti}, c_s^{ti}),
\end{equation*}
and implement the training as
\begin{equation*}
    \min_{\theta_t} \ \mathcal{L}_t^t.
\end{equation*}
Please note that there is no impact of $F_t(\phi_t)$ on $\mathcal{L}_t^t$ in $M_t$ because the residual adaptation block offers an identity channel for the target features. By doing this, we can eliminate the disturbance between modules and maintain the discriminative features as much as possible.

\subsubsection{Adaptation training of $F_t$}
Next, we need to align the source features to the target domain in $M_t$ to match its working condition. Similar as $F_s$, we minimize the $MMD_t$ defined in \ref{mmdt} to train $\phi_t$ as:
\begin{equation*}
    \min_{\phi_t} \ MMD_t.
\end{equation*}

\subsubsection{Supervised training of $C_t$ based on source samples}
After the alignment, the transformed source features now have the same distribution as the target domain. Then we utilize reliable label information of transformed source features to help the learning of discriminative structure in $C_t$. In this step, we only consider the source samples to implement a supervised learning process. Denote the logits for $x_s^i$ as $c_t^{si}$ as Equation \ref{logitsts}, we define the target-model source-sample loss as
\begin{equation*}
    \mathcal{L}_t^s = \sum_{i=1} ^{n_s} l(c_t^{si}, y_s^i).
\end{equation*}
Different from previous steps, $\mathcal{L}_t^s$ is affected by $\phi_t$, $\theta_t$ together. In this step, we only update $\theta_t$ and freeze $\phi_t$ to maintain the alignment and match $C_t$ to the updated $F_t$. In other words, we implement the training as
\begin{equation*}
     \min_{\theta_t} \ \mathcal{L}_t^s.
\end{equation*}

\subsubsection{Agreement feedback for $F_s$}
Now we complete the information flow from $M_s$ to $M_t$. Recall that our ultimate goal is to train accurate domain-wise models for both domains, under which these two models should reach the same decision for source and target samples. Since we have already considered the source labels during the training, we define an agreement loss $L_{s,t}^t$ to check the logits agreement between two models on target samples as follows:
\begin{equation*}
    \mathcal{L}_{s,t}^s = \sum_{i=1} ^{n_t} l(c_t^{ti}, c_s^{ti}).
\end{equation*}
We use the agreement loss $L_{s,t}^t$ to feed the information flow back to the source domain and only update the residual adaptation block $F_s$ as
\begin{equation*}
    \min_{\phi_s} \ \mathcal{L}_{s,t}^t.
\end{equation*}
Now we complement one epoch of the overall interactive training.

\subsection{Overall objectives}
In summary, the overall objective of the IRDAN can be defined as:
\begin{equation}\label{overallobj}
\begin{aligned}
\min_{\epsilon_s, \epsilon_t, \phi_s, \phi_t, \theta_s, \theta_t} L_s^{con} &+ L_t^{con} 
+ L_s^s + L_t^t + L_t^s \\ &+ MMD_s + MMD_t + L_{s,t}^t
\end{aligned}
\end{equation}

Even though the objective Equation \ref{overallobj} looks complex, we decompose it into several steps and train only one module in each step to maintain the overall stability and avoid the DAP problem. More importantly, in this process, we transfer the information from the module with the highest confidence (i.e., $C_s$) to the target models to make the model as reliable as impossible. We also formulate a feedback information flow to train these two models interactively and construct reliable training. After several epochs of training, all these losses should converge to small values. 

\begin{algorithm}
\caption{ Interactive training for IRDAN }
\begin{algorithmic}\label{alg}
\STATE \textbf{Input:} Labeled source samples $\{(x^i_s,y^i_s)\}_{i=1}^{n_s}$ and unlabeled target samples $\{x^i_t\}_{i=1}^{n_t}$, maximum epochs $n$, desired reward $V^*$ and initialized $\epsilon_s, \epsilon_t, \phi_s, \phi_t, \theta_s, \theta_t$.
\vspace{2mm}
\STATE $\epsilon_s \leftarrow \argminA \mathcal{L}_s^{con}$
\vspace{1mm}
\STATE $\epsilon_t \leftarrow \argminA \mathcal{L}_t^{con}$
\vspace{1mm}
\WHILE{$i \leq n$}
\vspace{1mm}
    \STATE $\theta_s \leftarrow \argminA\mathcal{L}_s^s,$
    \vspace{1mm}
    \IF{$V \geq V^*$}
    \vspace{1mm}
        \STATE \textbf{BREAK}
        \vspace{1mm}
    \ENDIF
    \vspace{1mm}
    \STATE $\phi_s \leftarrow \argminA MMD_s$
    \vspace{1mm}
    \STATE $\theta_t \leftarrow\argminA \mathcal{L}_t^t$
    \vspace{1mm}
    \STATE $\phi_t \leftarrow \argminA MMD_t$
    \vspace{1mm}
    \STATE $\theta_t \leftarrow \argminA \mathcal{L}_t^s$
    \vspace{1mm}
    \STATE $\phi_s \leftarrow \argminA \mathcal{L}_{s,t}^t$
    \vspace{1mm}

\ENDWHILE
\vspace{2mm}
\STATE \textbf{Return:} Updated $\epsilon_s, \epsilon_t, \phi_s, \phi_t, \theta_s, \theta_t$
\end{algorithmic}
\end{algorithm}

\subsection{Stopping criterion}
An essential problem for the deep learning-based models is ``when should we stop the training?'' because the training loss is not reliable enough to determine the best model. In ordinal settings, researchers use the validation set to monitor the validation accuracy of the model and choose the one with the highest validation accuracy. In the unsupervised domain adaptation scenarios, however, we do not have labeled target samples as the validation set. As a result, when should we stop the training for domain adaptation networks?

To the best of our knowledge, there is no related discussion about this problem. Users generally choose the model with the lowest training loss. However, the training loss is the overall objective and the source of gradients so it decreases in general. On the contrary, the accuracy is not always guaranteed to increase during the training process. As a result, this naive criterion is not reliable as we will show in Section \ref{indepth}.

In our IRDAN framework, we construct a reliable stopping criterion for the interactive training process based on the agreement between the two models. In domain adaptation scenarios, we care more about the decisions of target samples so we define the agreement between two models on the target samples as the reward $V$ as follows:
\begin{equation}\label{value}
    V = \frac{1}{n_t}\sum_{i=1}^{n_t}\mathbb{I}(M_t(x_t^i)=M_s(x_t^i)),
\end{equation}
where $\mathbb{I}(\cdot)$ represents the indicator function. $V$ actually represents the agreement percentage of two models on the target samples. Please note that this reward function is different from the agreement loss function $L_{s,t}^s$ because we can calculate $V$ at any status during the training but $L_{s,t}^s$ is only defined in a specific step.

We use $V$ to measure the convergence of the interactive framework: the higher the $V$, the convergent the whole training. Ideally, $V$ approximates $1$ if the whole training converges. In applications, we can set a desired reward $V^*$ and stop the training when $V$ reaches this value. Based on our experiments, we measure $V$ after step 1 to monitor the convergence of the whole training. Algorithm \ref{alg} shows the detailed training steps of IRDAN.

\section{Case study}\label{Casestudy}
In this section, we construct several PDA tasks in two cases (i.e., the CWRU dataset and the Paderborn dataset) to test the proposed method. We compare the proposed method with prevailing DA or PDA approaches: MCD \cite{saito2018maximum}, DANN \cite{ganin2016domain}, SAN \cite{cao2018partial1}, IWAN \cite{zhang2018importance} and DRCN \cite{li2020deep} to show the superiority of the proposed method. The experiments of CWRU dataset are included in Section II of Supplementary.

\subsection{Dataset and tasks}\label{data}
%\subsubsection{CWRU dataset}
%The mechanical rolling bearing dataset provided by Case Western Reserve University includes vibration signals collected from different positions of the bearing \cite{smith2015rolling}, i.e., fan end and drive end, and different working speeds, i.e., 1730 rpm, 1750 rpm, 1772 rpm and 1797 rpm. In this experiment, we choose the data collected on the drive end with a sampling frequency of 12kHz and consider two working speeds with the largest differences, i.e., 1730 rpm and 1797 rpm, to construct the domain adaptation scenario, i.e., 1797 rpm $\xrightarrow{}$ 1730 rpm.  

The Paderborn dataset is a large-scale collection of vibration signals collected from the rolling bearing test module \cite{lessmeier2016condition}. These damages are
classified into one of three categories: normal condition (N),
OR fault (O), and IR fault (I).  According to different rotational speed, load torque,
and radial force, the Paderborn dataset can be divided into
four domains: N15\_M07\_F10 (570), N09\_M07\_F10 (970),
N15\_M01\_F10 (510), and N15\_M07\_F04 (574). We construct twelve partial domain adaptation tasks as shown in Table \ref{tab:settings2}. In each domain, we have 1000 samples available for each category with every sample consisting of 1024 data points with no overlap.

    \begin{table}[ht]
        \centering
        \caption{PDA tasks for the Paderborn dataset}
        \begin{tabular}{cccccc}
        \toprule
             Task & Scenario & Target &Task & Scenario & Target \\
        
        \midrule
              B1 & 574$\xrightarrow{}$ 570  & N, I & B7 & 570$\xrightarrow{}$ 574  & I\\

        \midrule
              B2 & 574$\xrightarrow{}$ 570  & I & B8 & 570$\xrightarrow{}$ 574  & O\\
         \midrule
              B3 & 574$\xrightarrow{}$ 510  & N, I & B9 & 510$\xrightarrow{}$ 570  & N, I\\
         \midrule
              B4 & 574$\xrightarrow{}$ 510  & N, O & B10 & 510$\xrightarrow{}$ 570  & I\\
         \midrule
              B5 & 570$\xrightarrow{}$ 510  & N, I & B11 & 510$\xrightarrow{}$ 574  & N, O\\
         \midrule
              B6 & 570$\xrightarrow{}$ 510  & I & B12 & 510$\xrightarrow{}$ 574  & O\\

        \bottomrule
        \end{tabular}
        \label{tab:settings2}
    \end{table}

\subsection{Implement details}

In this work, we first transform the raw vibration signals with a time series structure into 2-D images based on wavelet transform following \cite{shao2018highly}. Based on the input dimension of the vibration signals, we transform each sample into an image in size $32\times32$, respectively. 

Table \ref{tab:structure} shows the detailed structure and corresponding main parameters of modules we use in all the experiments. The IRDAN uses the $G_s$, $G_t$, $F_s$, $F_t$, $C_s$, and $C_t$. In the pre-training stage of IRDAN, we train $G_s$, $G_t$ separately with epoch $=20$ and temperature $= 0.5$. In the interactive learning stage, we set epoch $=10$, and in each epoch, the number of iterations for every step is also set to $10$. In other words, in one epoch, each step includes $10$ iterations and each epoch includes $6\times10$ iterations for all six steps. As a result, we implement $10\times10$ iterations for each interactive learning step included in Section \ref{interactivelearning}. We set the learning rate $10^{-3}$ for all these steps and choose the model with the highest reward $V$ as our final model.

As for the other adversarial learning-based benchmarks, we use the same structure of $G_s$ or $G_t$ as the basic feature extractor denoted as $G$. Different from the IRDAN, we train $G$ based on source and target samples together based on the training strategies of these methods. In order to implement a fair comparison, we combine the structure of $F$ and $C$ in IRDAN to construct the classifier of adversarial learning-based benchmarks. We use the same learning rate and number of epochs as the IRDAN and choose the model with the least loss as the final model.

    \begin{table}[ht]
        \centering
        \caption{Implementation details}
        \begin{tabular}{ccc}
        \toprule
             Modules &  Layers & Parameters\\
        
        \midrule
              \multirow{7}*{$G_s$, $G_t$} &  Conv1d & $3\times3\times 32$\\
                &  BN, ReLU, Maxpooling & \\
               &  Conv1d & $3\times3\times 64$\\
                &  BN, ReLU, Maxpooling & \\
               &  Conv1d & $3\times3\times 128$\\
                &  BN, ReLU, Maxpooling & \\
                &  Flatten, FC & Output: 512\\ 
        \midrule
              \multirow{5}*{$F_s$, $F_t$} &  FC, BN, ReLU, Dropout & Output: 256\\
               &  FC, BN, ReLU, Dropout & Output: 128\\
               &  FC, BN, ReLU, Dropout & Output: 256\\
               &  FC, BN, ReLU, Dropout & Output: 512\\
               &  Identity channel & Output: 512\\
        \midrule
              \multirow{3}*{$C_s$, $C_t$} &  FC, BN, ReLU, Dropout & Output: 128\\
               &  FC, BN, ReLU, Dropout & Output: 64\\
               &  FC, BN, softmax, Dropout & Output: 3\\
         \midrule
              \multirow{2}*{\makecell[c]{Domain \\discriminator}} &  FC, BN, ReLU, Dropout & Output: 100\\
               &  FC, BN, softmax & Output: 2\\
        \midrule
              \multirow{5}*{Classifier} & FC, BN, ReLU, Dropout & Output: 256\\
              &  FC, BN, ReLU, Dropout & Output: 128\\
               &  FC, BN, ReLU, Dropout & Output: 64\\
               &  FC, BN, softmax, Dropout & Output: 3\\
        \bottomrule
        \end{tabular}
        \label{tab:structure}
    \end{table}

\subsection{Results analysis}
Table \ref{tab:puacc} shows the resulting accuracy of the seven methods. One can see that PDA-oriented approaches SAN, IWAN, and DRCN work better than the traditional DA approaches MCD and DANN, proving that the partial problem indeed deteriorates the performance of traditional DA approaches. Among these PDA approaches, the sample reweighting-based PDA approaches SAN and IWAN work much worse than DRCN which adds a residual adaptation block additionally, proving the efficiency of residual adaptation block. One can see that the IRDAN achieves around $4\%$ higher average accuracy on the basis of DRCN. The main reason is two-fold: Firstly, the proposed IRDAN constructs two domain-wise models with the residual domain adaptation blocks and proposes an interactive learning framework to train these modules one by one to avoid the prevailing ADP problem; Secondly, the IRDAN does not rule out any source samples based on the reweighting strategy to solve the PDA challenge and utilize the originally provided information as much as possible.

%Table \ref{tab:puacc} shows the accuracy of the seven methods for the Paderborn dataset. Different from the CWRU case, the IWAN performs slightly better than SAN. The possible reason may be that the fewer categories in the Paderborn dataset need fewer samples to learn the classification boundary. 

\begin{table}[ht]
    \centering
    \caption{Diagnosis accuracy of the seven methods for the Paderborn dataset}
    \begin{tabular}{ccccccccccccc}
    \toprule
         Task & MCD & DANN & SAN & IWAN & DRCN & IRDAN \\
    \midrule
    B1 & 86.35 & 87.35 & 85.75 & 86.35 & 88.50 & 94.50 \\
    B2 & 68.00 & 66.00 & 75.00 & 68.00 & 92.25 & 95.30 \\
    B3 & 91.35 & 91.60 & 88.65 & 92.30 & 95.25 & 97.05 \\
    B4 & 84.80 & 89.45 & 88.45 & 90.45 & 90.65 & 95.60 \\
    B5 & 91.50 & 89.50 & 83.50 & 90.25 & 89.25 & 96.00 \\
    B6 & 78.45 & 75.05 & 77.05 & 80.05 & 94.20 & 95.10 \\
    B7 & 83.85 & 82.80 & 80.80 & 83.15 & 91.55 & 96.50 \\
    B8 & 64.30 & 65.05 & 74.45 & 66.25 & 88.40 & 96.40 \\
    B9 & 83.75 & 85.70 & 82.75 & 85.45 & 89.85 & 92.50 \\
    B10 & 59.35 & 56.50 & 52.55 & 60.55 & 85.95 & 88.90 \\
    B11 & 92.35 & 94.05 & 87.45 & 94.25 & 94.35 & 97.75 \\
    B12 & 68.75 & 65.05 & 76.75 & 70.45 & 97.15 & 99.10 \\
    \midrule
    Avg. & 79.40 & 79.01 & 79.26 & 80.63 & 91.45& 95.39\\
    \bottomrule
    \end{tabular}
    \label{tab:puacc}
\end{table}

\subsection{Comparison of stopping criteria}\label{indepth}
\begin{figure*}[ht]
    \centering
    \includegraphics[width=\textwidth]{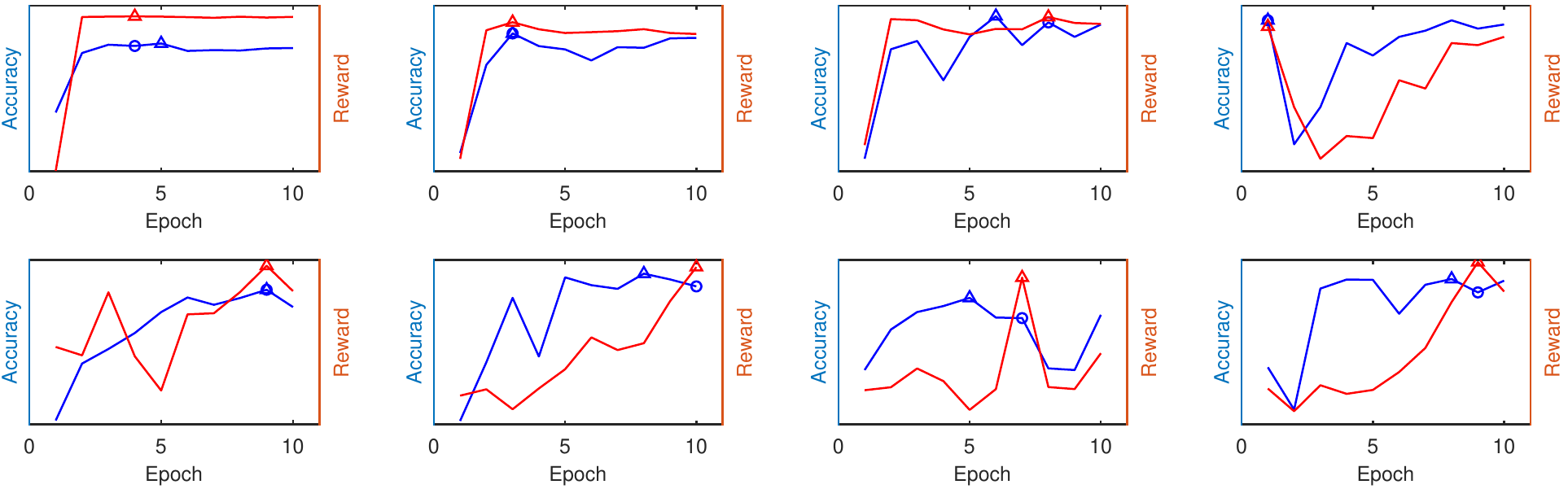}
    \caption{the correlation between the accuracy (blue lines) and the reward (red lines) for the proposed IRDAN and DANN. The four subfigures on the top show the accuracy and reward of IRDAN at each epoch on tasks B5, B9, B11 and B12 for the Paderborn dataset while the subfigures on the bottom represent the accuracy and reward of DANN for the same tasks. Triangles on the red lines note the maxima of the reward and the circles on the blue lines show the selected models based on the reward. Triangles on the blue lines show the real best accuracy.}
    \label{fig:accloss}
\end{figure*}

We compare the correlation between the accuracy (blue lines) and the reward (red lines) for the proposed IRDAN and DANN in Figure \ref{fig:accloss} to show the efficiency of our stopping criterion. The four subfigures on the top show the accuracy and reward of IRDAN at each epoch on tasks B5, B9, B11 and B12 for the Paderborn dataset while the subfigures on the bottom represent the accuracy and reward of DANN for the same tasks. Note that loss is inversely proportional to accuracy for DANN in general. To observe the difference between it and IRDAN (whose accuracy and reward are directly proportional) in a more intuitive way, we use the reciprocal of the loss as the reward for DANN. One can see the trend of blue lines is roughly similar to red lines for the IRDAN while the trends of the two lines do not coincide well for DANN. 

In most existing works, researchers generally choose the ``best'' performance calculated by the target labels as evidence to show the superiority of their works. However, we cannot get this performance on the target domain in fact due to the lack of target labels. As a result, we need a stopping criterion to choose the model in real-world applications. In order to prove the efficiency
of the proposed stopping criterion, we mark the maxima of the reward using triangles for the red lines. Then we note the corresponding accuracy on the blue lines using circles based on the same epoch guided by the red triangles. This process imitates how we choose the best model in real-world applications for IRDAN and DANN: select the model with the highest reward. In other words, the circles represent the real performance of IRDAN and DANN in the applications. We also mark the true best model with the highest accuracy on the target samples using triangles on the blue lines. If the difference between the blue circles and blue triangles is small, the corresponding stopping criterion is efficient. It is apparent that such differences for the subfigures on the top are much less than those on the bottom, which proves the efficiency of our stopping criterion. 

\section{Conclusion}
This work proposes a novel framework called the residual interactive domain adaptation networks (IRDAN) to solve the partial domain adaptation challenge in industrial fault diagnosis tasks. Inspired by the talents of ordinary classification settings with identical distribution, we construct domain-wise models for the corresponding domain distribution to solve the partial challenge from a new direction. Then we equip each domain-wise model with a residual domain adaptation (RDA) block to avoid the adaptation-discrimination paradox prevailing in the adversarial learning-based PDA methods. Additionally, we formulate a confident information flow via an interactive learning strategy to train the modules of IRDAN one by one and avoid interference between them. We also set up a reliable stopping criterion for the IRDAN to choose the best model for users in real-world applications. The experiments have proven the superiority of the proposed IRDAN.

\bibliographystyle{Bibliography/IEEEtranTIE}
\bibliography{Bibliography/IEEEabrv,Bibliography/BIB_xx-TIE-xxxx}\ %IEEEabrv instead of IEEEfull

%\bibliography{ref}
%\bibliographystyle{IEEEtran}
\ %IEEEabrv instead of IEEEfull

\end{document}